\documentclass{ecai}
\usepackage{times}
\usepackage{graphicx}
\usepackage{latexsym}
\usepackage{hyperref}

\usepackage{multirow}
\usepackage{algorithm}
\usepackage[noend]{algpseudocode}
\usepackage{algorithmicx}
\usepackage{amssymb}
\usepackage{amsmath}

\ecaisubmission

\makeatletter
\def\blfootnote{\gdef\@thefnmark{}\@footnotetext}
\makeatother

\begin{document}

\title{Neural Large Neighborhood Search for the Capacitated Vehicle Routing Problem}

\author{Andr\'{e} Hottung \and Kevin Tierney\institute{Bielefeld University,
Germany, email: \{andre.hottung, kevin.tierney\}@uni-bielefeld.de} }

\maketitle
\blfootnote{Accepted at ECAI 2020. \url{https://doi.org/10.3233/FAIA200124}}
\bibliographystyle{abbrv}

\begin{abstract}
Learning how to automatically solve optimization problems has the potential to provide the next big leap in optimization technology.
The performance of automatically learned heuristics on routing problems has been steadily improving in recent years, but approaches based purely on machine learning are still outperformed by state-of-the-art optimization methods. To close this performance gap, we propose a novel large neighborhood search (LNS) framework for vehicle routing that integrates learned heuristics for generating new solutions.
The learning mechanism is based on a deep neural network with an attention mechanism and has been especially designed to be integrated into an LNS search setting. 
We evaluate our approach on the capacitated vehicle routing problem (CVRP) and the split delivery vehicle routing problem (SDVRP). On CVRP instances with up to 297 customers, our approach significantly outperforms an LNS that uses only handcrafted heuristics and a well-known heuristic from the literature. Furthermore, we show for the CVRP and the SDVRP that our approach surpasses the performance of existing machine learning approaches and comes close to the performance of state-of-the-art optimization approaches.
\end{abstract}

\section{INTRODUCTION}
Recent advances in the field of machine learning have allowed neural networks to learn how to perform a wide variety of tasks. In the area of optimization, there has been growing interest in using deep reinforcement learning to automatically learn heuristics for optimization problems. Especially for practical applications, the automated generation of \textit{good enough} heuristics is of great interest, because the costs (in terms of labor) associated with developing hand-crafted heuristics is not always worth the performance gains.
While there has been success in automatically learning heuristics that outperform state-of-the-art techniques on some problems, e.g., container pre-marshalling~\cite{DLTS}, many problems have proven extremely difficult for learned heuristics.

Routing problems, such as the traveling salesman problem (TSP) and the vehicle routing problem (VRP), are among the most widely solved optimization problems in practice. However, the highly variable size of the problem representations have long made an application of machine learning based methods difficult.
The introduction of advanced deep learning model architectures, such as pointer networks~\cite{vinyals2015pointer}, have now made it possible to handle variable length sequences of input data, making the application of machine learning methods to routing problems practical. 

Several approaches have been proposed that learn how to solve routing problems, but even the most recent methods do not yet outperform state-of-the-art optimization techniques. This performance gap can be partially explained by the simple search strategies of machine learning based methods, e.g., sampling \cite{kool2018attention} or beam search (BS) \cite{nazari2018reinforcement}. To address this, we propose to integrate learned heuristics into a higher level metaheuristic. We develop a large neighborhood search (LNS) metaheuristic for the VRP that learns heuristics for repairing incomplete solutions and employs them to guide a search through the solution space. Note that the developed metaheuristic does not contain any components that have been specifically developed for a certain routing problem. As in previous approaches, our method learns the complex heuristics needed to solve a class of problem instances on its own.

In this work, we focus on the capacitated vehicle routing problem (CVRP) and the related split delivery vehicle routing problem (SDVRP). The CVRP was introduced by \cite{dantzig1959truck} and is one of the most well researched problems in the optimization literature. A problem instance is described by a complete undirected graph $G = (V , E)$, with $V = \{v_0,...,v_k\}$. Node $v_0$ represents the depot, and all other nodes represent the customers. Each customer has a demand $d_i$ and the costs of traveling from node $v_i$ to $v_j$ are given by $\bar{c}_{ij}$. A fleet of vehicles is available that each have capacity $Q$. The task is to find a set of routes (all starting and ending at the depot) with minimal cost so that the demand of all customers is fulfilled and each customer is visited by exactly one vehicle. The SDVRP differs from the CVRP only by the fact that a customer can be visited by multiple vehicles (i.e., a delivery can be split into multiple deliveries). We consider the versions of the CVRP and SDVRP where the distance matrix obeys the triangle inequality.

We propose a new approach, called neural large neighborhood search (NLNS) that integrates learned heuristics in a sophisticated high level search procedure. NLNS is based on large neighborhood search (LNS), a metaheuristic that explores the space of all possible solutions by iteratively applying destroy and repair operators to a starting solution. We implement two simple destroy procedures that can be applied to any routing problem. The significantly more complex task of repairing a destroyed (i.e. incomplete) solution is left to a deep neural network that is trained via policy gradient reinforcement learning. We choose LNS as the foundation for our approach for several reasons. First, LNS offers a simple framework to learn a neighborhood function via its destroy and repair paradigm. Second, the complexity of the repair problem is mostly independent of the instance size, allowing us to tackle much larger problem instances than previous approaches. Finally, LNS has been successfully applied to a large number of optimization problems, including many routing problems (e.g., \cite{ropke2006adaptive}). We evaluate NLNS on a diverse set of CVRP and SDVRP instances. We show that NLNS offers a significant improvement over an LNS using a handcrafted repair operator from the literature. Furthermore,  NLNS significantly outperforms existing machine learning approaches and comes close to or matches the performance of state-of-the-art optimization approaches.  

The contributions of this work are as follows: 
\begin{enumerate}
    \item We develop an LNS that relies on learned heuristics to explore the neighborhood of a solution. Our approach has been specifically adapted to benefit from the parallel computing capabilities of a GPU.
    \item We propose a new neural network model for the VRP with an attention mechanism for completing incomplete solutions. 
\end{enumerate}

This paper is organized as follows. First, we discuss related work that uses machine learning based methods to tackle routing problems. We then introduce the NLNS approach along with the new neural network model. Finally, we evaluate NLNS on CVRP and SDVRP instances and compare it against state-of-the-art optimization and machine learning methods.  

\section{RELATED WORK}
The first application of neural networks to combinatorial optimization problems is described by \cite{hopfield1985neural}, in which a Hopfield network is used to compute solutions for the TSP. Despite this promising early application, learning-based approaches have only recently become a more serious contender to traditional optimization approaches thanks to the introduction of new neural network architectures. The pointer network introduced in \cite{vinyals2015pointer} uses an attention mechanism to learn the conditional probability of a permutation of a given input sequence (e.g., a permutation of the customers of a TSP instance). The authors train their model to solve TSP instances of up to 50 customers using supervised learning. During inference, a beam search is used and they report promising results for TSP instances with up to 30 customers. The idea is extended by \cite{bello2016neural}, in which a pointer network is trained using actor-critic reinforcement learning. The authors report improved performance over a supervised learning based approach for TSP instances with 50 and 100 customers.

A number of other approaches have been proposed for the TSP. In \cite{khalil2017learning}, a graph embedding network is proposed that learns a greedy heuristic that incrementally constructs a solution.  A graph neural network is used in \cite{nowak2017note} together with a beam search (with beam width size 40) to generate solutions for the metric TSP. The network is trained via supervised learning and the reported performance is slightly worse than the performance of the pointer model. Another supervised learning based approach is proposed by \cite{kaempfer2018learning} for the multiple TSP. The authors use a permutation invariant pooling network in combination with beam search to generate solutions. A graph attention network \cite{velivckovic2017graph} is trained in \cite{deudon2018learning} via reinforcement learning to build solutions for the TSP that are improved in a subsequent step via a 2-OPT local search. In \cite{joshi2019efficient}, a graph convolutional network \cite{bresson2017residual} is used to generate an edge adjacency matrix describing the probabilities of edges occurring in a TSP solution. A post-hoc beam search is used to convert the edge probabilities to a valid solution.

To date, only a few deep learning based approaches exist that consider the VRP. In \cite{nazari2018reinforcement}, a model is proposed that uses an attention mechanism and a recurrent neural network (RNN) decoder to build solutions for the CVRP and the SDVRP. In contrast to the pointer model, the proposed model is invariant to the order of the inputs (i.e., the order of the customers). Furthermore, the model can efficiently handle dynamic input elements (e.g., customer demands that change once a customer has been visited). The authors use an actor-critic reinforcement learning approach similar to \cite{bello2016neural} and search for good solutions using a beam search with a beam width of up to 10. In contrast to NLNS, solutions are built using a sequential approach, where the customer that was visited in the last step is always connected to another customer in the next step. This allows the RNN decoder to learn a decoding strategy that takes the output of previous decoding steps into account. In NLNS, an incomplete solution is repaired and the current state of the solution does not solely depend on the decision made in previous repair steps. Hence, we do not use an RNN in our model.

A graph attention network similar to \cite{deudon2018learning} is used in \cite{kool2018attention} and trained via reinforcement learning to generate solutions for different routing problems including the TSP, the CVRP, and the SDVRP. The authors train their model using policy gradient reinforcement learning with a baseline based on a deterministic greedy rollout. In contrast to our approach, the graph attention network uses a complex attention-based encoder that creates an embedding of a complete instance that is then used during the solution generation process. Our model only considers the parts of an instance that are relevant to repair a destroyed solution and does not use a separate encoder. The search procedure employed in \cite{kool2018attention} samples multiple solutions for a given instance in parallel and returns the best one.

In parallel work, a neural network is used in an LNS setting to solve a routing problem encountered in vehicle ride hailing \cite{syed2019neural}. The employed network is trained via supervised learning and its input data is composed of complex and domain specific features (e.g., regret value). In contrast, NLNS aims at automating the generation of heuristics in a way that requires no deep problem or optimization knowledge. 

A large number of optimization approaches exist that use machine learning components, e.g., algorithm selection approaches that learn to select an algorithm out of a portfolio of options or approaches that learn to adjust search parameters during the search (e.g., \cite{ansotegui2018hyper}). We do not discuss these approaches here, but refer readers to \cite{kotthoff2016algorithm} and \cite{bengio2018machine}.

Machine learning techniques are often also used in hyper-heuristics, which also aim at automating the design of metaheuristic methods. Hyper-heuristics achieve this by choosing or combining existing heuristics or heuristic components. For example, in~\cite{tyasnurita2017learning} a hyper-heuristic for a VRP variant is proposed that learns how to apply low-level heuristics during the search process.    

\section{NEURAL LARGE NEIGHBORHOOD SEARCH}\label{sec:nlns}
NLNS is an extension to the LNS metaheuristic that automates the complex task of developing repair operators using reinforcement learning. NLNS has been designed to make full use of the parallel computing capabilities of a GPU and supports two search modes: \textit{batch search}, in which a batch of instances is solved simultaneously, and \textit{single instance search}, in which a single instance is solved using a parallel search.

LNS is a metaheuristic in which the neighborhood of a solution is implicitly defined by a repair and a destroy operator and was first introduced by Shaw~\cite{shaw1998using}. Let $\pi$ be an initial solution for a given problem instance. LNS generates a neighboring solution $\pi'$ of $\pi$  by applying a destroy operator followed by a repair operator. The destroy operator deletes parts of the solution, e.g., by removing tours from a solution to the CVRP, resulting in an infeasible solution. It usually contains stochastic elements so that different parts of a solution are deleted when the same solution is destroyed multiple times. The repair operator then fixes the destroyed solution, e.g., for the CVRP by completing incomplete tours or assigning unassigned customers to existing tours, thus creating a feasible solution $\pi'$. An acceptance criterion such as the Metropolis criterion from simulated annealing~\cite{kirkpatrick1983optimization} is used to determine whether the new solution should be \textit{accepted}, i.e., whether $\pi'$ should replace $\pi$. After updating $\pi$ (or not), the search continues until a termination criteria is reached.

The performance of LNS heavily depends on the quality of the destroy and repair operators developed by domain experts. While even simple destroy operators that destruct parts of a solution purely at random can lead to good solutions, the repair operators often require the implementation of a sophisticated heuristic. In contrast to LNS, NLNS uses multiple destroy and repair operators (as in adaptive large neighborhood search \cite{ropke2006adaptive}). Each repair operator $o^R$ corresponds to a learned parameterization $\theta$ of our proposed neural network model that repairs a solution in a sequential repair process. During the training of a repair operator, the corresponding model is repeatedly presented incomplete solutions (to instances sampled from a certain distribution) that have been destroyed with a particular destroy operator~$o^D$. The objective of the training is to adjust the model parameters~$\theta$ so that the model constructs high quality solutions. By presenting the model only instances from a certain distribution, it is able to learn repair operators adapted to the characteristics of specific instance classes and destroy operators. This capability is of particular importance for problems where instances with similar characteristics are frequently solved in practice, as is the case for VRP variants. This also means that NLNS can be retrained in the case that the characteristics of the encountered instances change, avoiding significant human resources needed to design new operators.  

The initial starting solution $\pi$ is generated by a simple greedy procedure. Starting from the depot a route is created by always adding the closest customer to a tour. If the demand of the closest customer exceeds the remaining load of the vehicle, the vehicle returns to the depot. This process is repeated until all customers have been visited. The result is a solution $\pi$ that consists of a set of tours all starting and ending at the depot. 

\subsection{Destroy operators}
A solution is destroyed using a destroy operator $o^D$ that specifies how many elements should be removed from a solution and how these elements should be selected. To destroy a solution, we only implement two simple destroy procedures that require no deep domain knowledge and can be applied to almost any routing problem. 
\begin{itemize}
    \item \textit{Point-based destroy} removes the customers closest to a randomly selected point from all tours of a solution.
    \item \textit{Tour-based destroy} removes all tours closest to a randomly selected point from a solution.
\end{itemize}
If a customer $v_j$ is removed from a tour $\{v_i,...,v_j,...,v_k\}$, three incomplete tours are created. The incomplete tour $\{v_i,...,v_{j-1}\}$ contains all vertices before customer $v_j$, the incomplete tour $\{v_j\}$ contains only customer $v_j$, and the tour $\{v_{j+1},...,v_{k}\}$ contains all vertices after customer $v_j$. If a complete tour consisting of $r$ customers is removed, $r$ incomplete tours consisting of only one customer each are created. Each destroy operator specifies how many customers  should be removed in percent (this is known as the \textit{degree of destruction}).

\subsection{Learning to repair solutions}
The problem of repairing a destroyed solution can be formulated as a reinforcement learning problem in which an agent (a learned model) interacts with an environment (an incomplete solution) over multiple discrete time steps. In each time step the model connects an incomplete tour with another incomplete tour or with the depot. The process is repeated until the solution only contains complete tours that do not exceed the vehicle capacity. In the following, we first explain the repair process on a concrete example. We then give a more formal description on how the model input is generated and how we formalize the reinforcement learning problem.

\begin{figure}[]
\centerline{
\includegraphics[width=0.95\columnwidth]{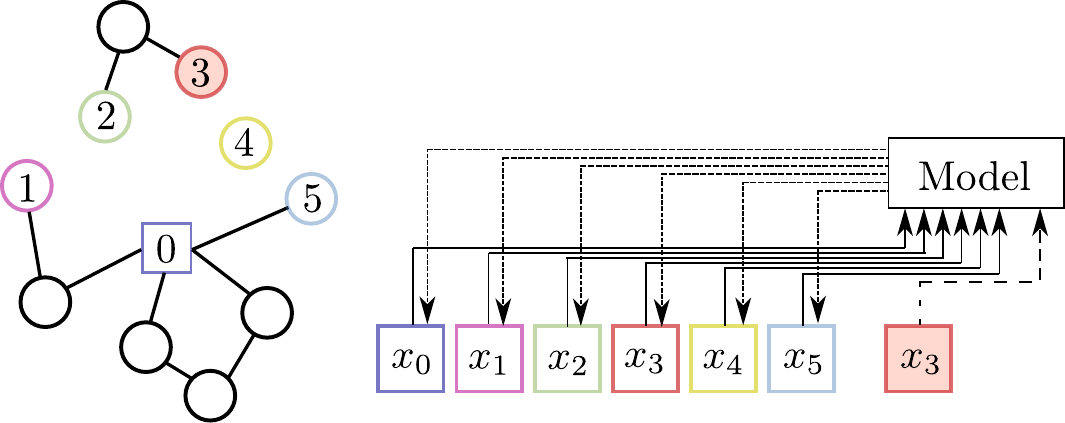}}
\caption{Incomplete solution and associated model input.}
\label{fig:repair_problem}
\end{figure}

Figure~\ref{fig:repair_problem} shows a destroyed solution and the corresponding model input. The solution contains four incomplete tours, with one incomplete tour consisting of only a single customer. The model receives a feature vector for each end of an incomplete tour that is not connected to the depot ($x_1,...,x_5)$. Each feature vector not only represents the node at the considered end of the tour, but also contains information on the whole incomplete tour, e.g., its unfulfilled demand. Additionally, the model receives a feature vector describing the depot ($x_0$). Furthermore, the model gets one end of a tour as a reference input. In this example, the reference input is $x_3$. The task of the model is to select where the associated tour end $3$ should be connected. The model does this by pointing to one of the inputs $x_0,...,x_5$. For example, if the model points to $x_5$, the tour end $3$ is connected to the tour end $5$. Note that the model is only getting information on the incomplete tours of a destroyed solution. Thus the size of the model input does not depend on the size of the instance (i.e., the number of customers), but on the degree of destruction. This allows us to apply NLNS to large instances with close to 300 customers.

The  model input for a destroyed solution $\pi$ is generated as follows. For each incomplete tour consisting of more than one customer, an input is created for each end that is not the depot (Step 1). Then, for each incomplete tour with only one node, a single input is generated (Step 2). Finally, an input for the depot is created (Step 3). Each input $x$ is a 4-dimensional feature vector represented by the tuple $\langle x^X, x^Y, x^D, x^S \rangle$. The values are calculated as follows. In the first case corresponding to Step 1, in which $x$ represents the end of an incomplete tour $\{v_i,...,v_j\}$, $x^X$ and $x^Y$ are the $x$ and $y$-coordinates of the node at the end of the tour under consideration and $x^D$ is the sum of the demands fulfilled by the tour $\{v_i,...,v_j\}$. The value $x^S$ is set to $3$ if the tour $\{v_i,...,v_j\}$ contains the depot and to $2$ otherwise. Corresponding to Step 2, if  $x$ represents a tour with a single node $\{v_i\}$, $x^X$ and $x^Y$ are the $x$ and $y$-coordinates of the node $v_i$ and $x^D$ is set to the fulfilled demand of the tour $\{v_i\}$\footnote{In the case of the CVRP, this is equal to $d_i$. In the case of the SDVRP, we compute the fulfilled demand as other tours may also include node $v_i$.}. The value $x^S$ is set to $1$. For an incomplete tour in Step 3, $x$ represents the depot $v_0$ and $x^X$ and $x^Y$ are the $x$-coordinate and the $y$-coordinate of the depot and $x^D$ and $x^S$ are set to $-1$. Before being fed into the network, the values $x^X$, $x^Y$, and $x^D$ are re-scaled to the interval $[0, 1]$  (except for the $x^D$ value of the depot, which is always $-1$). In the following, we describe how the model inputs are used in the sequential repair process.

Let $\pi_0$ be the incomplete solution that is generated by applying a destroy operator to a complete solution. In each repair time step $t$ the model is given the incomplete solution $\pi_t$ in the form of the tuple $(X_t, f_t)$, where $X_t$ contains all inputs generated using the procedure described above at time step $t$, and $f_t$ describes the reference input. The model then outputs a probability distribution over all actions. That is, the model defines a parameterized stochastic policy $p_\theta(a_t|\pi_t)$ with $\theta$ representing the parameters of the model. Each action $a_t$ corresponds to connecting the end of a tour represented by $f_t$ to one of the elements in $X_t$ and leads to the (incomplete) solution $\pi_{t+1}$. This process is repeated until a solution $\pi_T$ is reached that does not have any incomplete tours.

The reference input $f_t$ is an element of $X_t \setminus{\{x_0} \}$ and is selected at random at time step $t=0$. At the following time steps, $f_t$ is set to the input representing the end of the tour that was connected to $f_{t-1}$ if the tour associated with $f_{t-1}$ is still incomplete. Otherwise, $f_t$ is selected at random as in the first iteration. 

We use a masking procedure that does not allow the model to perform actions that would lead to infeasible solutions (e.g., creating a tour where the demand carried is higher than the vehicle capacity). We also do not allow the model to select the action that corresponds to connecting $f_t$ with itself. The masking is implemented by setting the log probability of forbidden actions to $0$.

\subsubsection{Model Architecture}
\begin{figure}[tb]
\centering
\includegraphics[width=0.9\columnwidth]{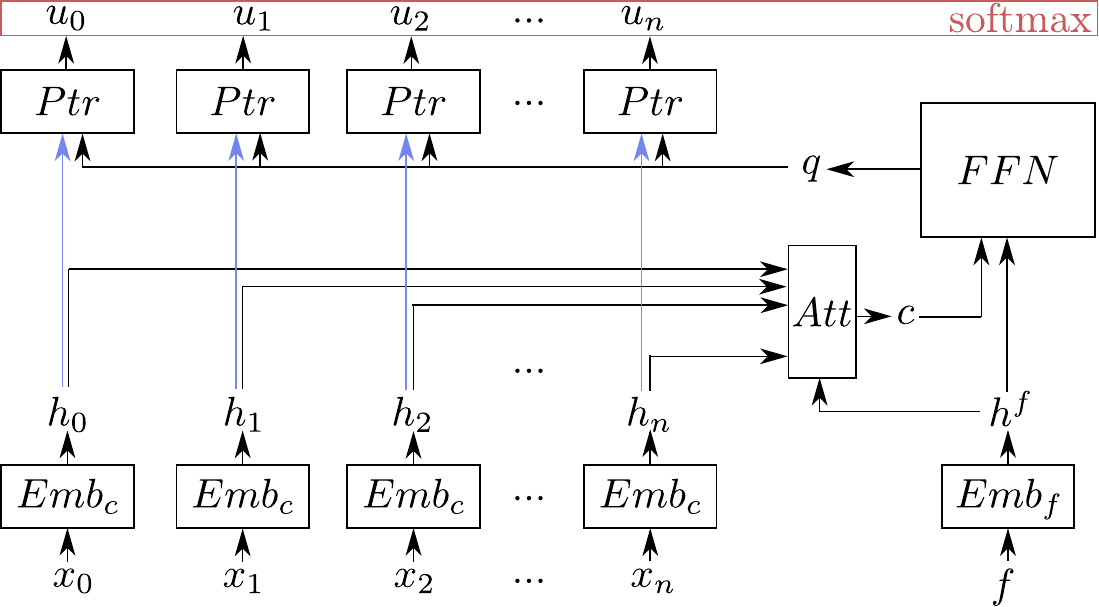}
\caption{Network architecture}
\label{att_network}
\end{figure}

The architecture of the model is shown in Figure~\ref{att_network}. The model is given the pair $(X_t, f_t)$ as input, which represents an incomplete solution $\pi_t$ at a time step $t$. For each of the inputs $x_i \in X_t$ an embedding $h_i$ is computed using the transformation $\mathit{Emb_c}$. $\mathit{Emb_c}$ consists of two linear transformations with a \textit{ReLU} activation in between\footnote{We noticed experimentally that the two layer transformation improves the performance over a single layer linear transformation, although other architectures may also be effective.}. It is applied to all inputs separately and identically. Both layers of $\mathit{Emb_c}$ have a dimensionality of $d_h$ (we set $d_h$ to 128 for all trained models). For the reference tour end representation $f_t$, an embedding $h^f$ is generated using the transformation $\mathit{Emb_f}$ that has the same structure as $\mathit{Emb_c}$, but uses different trainable parameters. All embeddings are used by the attention layer $\mathit{Att}$ to compute a single $d_h$-dimensional context vector $c$ that describes all relevant embeddings $h_0,\dots,h_n$. Which inputs are relevant is determined based on $h^f$. For example, the vector $c$ might contain mainly information on the inputs representing tour end points that are close to the tour end point represented by $f$. To compute the context vector $c$, first the $n$-dimensional alignment vector $\bar{a}$ is computed that describes the relevance of each input:
\begin{equation}
    \bar{a} = \mathit{softmax}(u^H_0,...,u^H_n),
\end{equation}
where
\begin{equation}
    u^H_i = z^A \tanh(W^A[h_i;h^f]).
\end{equation}

Here, $z^A$ is a vector and $W^A$ is a matrix with trainable parameters and ``;'' is used to describe the  concatenation of two vectors. Based on the alignment vector $\bar{a}$, the context vector $c$ is generated:
\begin{equation}
    c = \sum_{i=0}^n \bar{a}_ih_i.
\end{equation}

The context vector $c$ encodes information on relevant inputs, with the relevance of each input $x_i \in X_t$ given by the alignment vector $\bar{a}$. The concatenation of the vector $c$ and the embedding of the reference input $h^f$ is then given to a fully connected feed-forward network with two layers (both using a \textit{ReLU} activation) that outputs a $d_h$ dimensional vector $q$. This single output vector $q$ is used together with each embedding $h_0,\dots,h_n$ to calculate the output distribution over all actions:
\begin{equation}
		p_\theta(a_t|\pi_t) = \mathit{softmax}(u_0,...,u_n),
\end{equation}
where
\begin{equation}
    u_i = z^B \tanh(h_i + q),
\end{equation}
and the vector $z^B$ and contains trainable parameters.

In contrast to the architecture proposed in \cite{nazari2018reinforcement}, we do not use an RNN-based decoder. Furthermore, we use an FFN that computes the vector $q$ that is used to select an action based on the context vector $c$ (describing the relevant input points) and the vector $h^f$ (describing the reference input $f$). In  \cite{nazari2018reinforcement} the alignment vector $a$ is computed using the output of the RNN and the context vector $c$ is used directly to to calculate the output distribution over all actions.    

\subsubsection{Model Training}
We train our proposed model to repair solutions that have been destroyed with a specific destroy operator $o^D$ via reinforcement learning. To this end, we define a loss that is used to measure the quality of the actions performed by the model. Let $\pi_0$ be a solution after a destroy operator has been applied and let $\pi_T$ be the repaired solution after the model performs the actions $a_0,\dots,a_{T-1}$. The loss $L(\pi_T|\pi_0)$ describes the cost of repairing a solution, which is the difference between the total tour length of the destroyed solution and the total tour length of the repaired solution. The objective of the training is to adjust the parameters $\theta$ of the model to minimize the expected loss when repairing $\pi_0$,
\begin{equation}
    J (\theta|\pi_0) = \mathbb{E}_{\pi_T \sim p_{\theta}(.|\pi_0)} L(\pi_T|\pi_0),
\end{equation}
where we decompose the probability of generating solution $\pi_T$ as
\begin{equation}
    p_{\theta}(\pi_T|\pi_0) = \prod_{t=0}^{T-1}{p_{\theta}(a_t|\pi_t)},
\end{equation}
similar to \cite{sutskever2014sequence}.

We use the REINFORCE approach proposed by \cite{williams1992simple} to calculate the gradient
\begin{equation}
    \nabla J (\theta|\pi_0) = \mathbb{E}_{\pi_T \sim p_{\theta}(.|\pi_0)} [(L(\pi_0, \pi_T)-b(\pi_0)) \nabla_{\theta} \log p_{\theta}(\pi_T|\pi_0)],
\end{equation}
with $b(\pi_0)$ being the baseline. 
A good baseline that estimates the costs of repairing a solution $J (\theta|\pi_0)$ correctly can stabilize the training process. Similar to \cite{bello2016neural}, we use a second model called a $critic$ to estimate the cost of repairing $\pi_0$. The critic is trained in alternation with the policy model to minimize the mean squared error between its prediction for the costs of repairing $\pi_0$ and the actual costs when using the most recently learned policy. The critic is given as input the values $X_0$ generated for $\pi_0$ as described above. The critic processes each input using a position-wise feed-forward network that outputs a continuous value for each input. The sum of all outputs is then the estimate of the repair costs $b(\pi_0)$.

\subsection{Search Guidance}
In contrast to existing, purely machine learning based methods for the VRP, NLNS implements a problem-independent high level search procedure that enables a more effective exploration of the search space. To allow for a comparison to state-of-the-art machine learning and optimization methods, we implement two different version of the search procedure, both of which are designed to exploit the parallel computing capabilities of a GPU. \textit{Batch search} quickly solves a set of different instances in parallel and \textit{single instance search} solves a single instance by applying destroy and repair operators to multiple solutions of the instance at the same time. Both approaches use a given set of learned repair operators during the search with each repair operator $o^R$ being specifically trained for a given destroy operator $o^D$. Since GPUs excel only at applying \textit{the same} operation in parallel, only one repair operator $o^R$ can be used at a time.

\subsubsection{Single Instance Search}
The single instance search is geared towards the standard search setting for the VRP in which a single instance must be solved within a given time budget. To avoid getting stuck in a local minimum, LNS generally uses a simulated annealing based acceptance criteria \cite{kirkpatrick1983optimization} that accepts worsening solutions based on a temperature value $T$.

Algorithm~\ref{alg:single} shows the single instance search. At the beginning of the search the incumbent solution $\pi^*$ is set to the given feasible solution $\pi$ and the temperature $T$ is set to the start temperature $T_s$. The main search component consists of two nested while loops. In each iteration of the outer loop a batch $B$ consisting of copies of the incumbent solution $\pi^*$ is created. The solutions in $B$ are then improved in the inner loop using a gradually decreasing temperature $T$ (i.e., worsening solutions are accepted with decreasing likelihood). At the end of the outer loop, $T$ is set to the reheating temperature $T_r$ to allow the search to escape from its local minimum at the beginning of the next execution of the inner loop. 

In the inner loop, all solutions in $B$ are first destroyed and then repaired using a destroy and repair operator pair that is selected uniformly at random\footnote{The mechanism of adaptive large neighborhood search (ALNS)~\cite{ropke2006adaptive} could also be used, but we saw experimentally that a uniform random approach works best.}. The best solution $\pi^b \in B$ is then either accepted as the current solution $\pi$ or rejected. Furthermore, the incumbent is updated if $\pi^b$ is better than $\pi^*$. Finally, the temperature is updated and the first $Z\%$ of elements in $B$ are set to the current solution $\pi$. The parameter $Z$ controls to what degree the search focuses on generating neighboring solutions to the current solution $\pi$. The $(1-Z)\%$ last solutions in $B$ evolve completely independent of all other solutions in $B$, improving the diversity of the search. The single instance search can be further parallelized by running multiple instantiations of the inner loop in parallel that asynchronously exchange the incumbent solution $\pi^*$ after each execution of the inner loop.

\begin{algorithm}
    \footnotesize
    \textbf{Input:} A feasible solution $\pi$, start temperature $T_s$, reheating temperature $T_r$, minimum temperature $T_m$, cooling schedule $\delta$, percentage $Z$ of batch that is reset to the current solution  \\
    \textbf{Output:} The best found feasible solution $\pi^*$\\
    \begin{algorithmic}[1]
        \Function{NLNS-Single}{$\pi, T_s, T_r, T_m, \delta$, Z}
             \State $\pi^* \gets \pi$
             \State $T \gets T_{s}$ \Comment{Set to start temperature}
             \While{search termination criteria not reached}
                 \State $B \gets \{\pi^*,...,\pi^*\}$ \Comment{Create a batch of copies of $\pi^*$}
                 \While{$T > T_m$}
                 \State $B \gets \textsc{Repair}(\textsc{Destroy}(B))$
                 \State $\pi^b \gets \mathop{\rm arg\,min}_{\pi \in B}\{\textsc{Cost}(\pi)\}$
                 \If{$\textsc{Accept}(\pi^b, \pi^*, T)$}
                     \State $\pi \gets \pi^b$
                 \EndIf
                 \If{$\textsc{Cost}(\pi^b) < \textsc{Cost}(\pi^*)$}
                     \State $\pi^* \gets \pi^b$
                 \EndIf
                 \State Set the first $Z$\% of elements in $B$ to $\pi$
                 \State $T \gets \textsc{Update}(T, \delta)$
             \EndWhile
             \State $T \gets T_{r}$ \Comment{Set to reheating temperature}
             \EndWhile
        \State \algorithmicreturn{} $\pi^*$
        \EndFunction
    \end{algorithmic}
    \caption{Single Instance Search\label{alg:single}}
\end{algorithm}

\subsubsection{Batch Search}
In batch search, multiple instances are solved in parallel using the following procedure. At first, an initial solution is generated for each instance using the greedy heuristic as previously described. This creates a batch of solutions $B$. In each of the following iterations of the search, a destroy and repair operator pair is selected and all solutions in B are destroyed and repaired using the pair. This creates one neighboring solution $\pi'$ for each solution $\pi \in B$. A neighboring solution $\pi'$ is only accepted if the cost of $\pi'$ is less than the cost of the current solution $\pi$. The search is continued until the termination criteria for the entire batch is reached, e.g., a fixed time limit. The search can be parallelized by splitting $B$ into multiple subsets that are then processed individually in parallel. We do not use simulated annealing as in the single instance search, because in batch search significantly fewer destroy and repair operations are performed \emph{per instance}, making a sophisticated acceptance criteria less effective.

As in adaptive large neighborhood search (ALNS) \cite{ropke2006adaptive}, we select a repair operator $o^R$ according its performance in previous iterations. To this end, we track the absolute improvement of the average costs of all solutions in $B$ using an exponential moving average. However, in contrast to ALNS, which selects destroy and repair operators at random according to a learned probability distribution, we always select the operator pair that offered the best performance in previous iterations. Since the performance of operators converges towards zero over the course of the search, this ensures that all operators are used (and evaluated) regularly. This straightforward adaption mechanism offered a similar performance to more complex implementations in preliminary experiments. 

\section{COMPUTATIONAL RESULTS}
We evaluate NLNS on several CVRP and SDVRP instance sets. We split our analysis in two parts. First, we evaluate the batch search and compare it to existing machine learning approaches. Then, we compare the single instance search to state-of-the-art optimization approaches. We also evaluate the contribution of the learning-based components of NLNS by replacing the learned repair operators with the handcrafted repair operator proposed in~\cite{christiaens2016fresh}.

In all experiments, NLNS is run on a single Nvidia Tesla V100 GPU and a Intel Xeon Silver 4114 CPU at 2.2 GHz, using all 10 cores. Each repair operator is trained on 250,000 batches of 256 instances using the Adam optimizer~\cite{kingma2014adam} and a static learning rate of $10^{-4}$. The time to train a repair operator ranges from several hours (e.g., for instances with 20 customers) to 4 days (for instances with close to 300 customers)\footnote{Our PyTorch implementation of NLNS is available at \url{https://github.com/ahottung/NLNS}}.

\subsection{Batch Search}
We compare the NLNS batch search to the attention model with sampling (AM) from \cite{kool2018attention} and the reinforcement learning approach with beam search (RL-BS) from \cite{nazari2018reinforcement} on CVRP and SDVRP instances with 20, 50 and 100 customers. For each of the six problem classes we consider, we define four destroy operators: two for each destroy procedure, each with a different degree of destruction, and train four corresponding repair operators.

For our comparison to the AM approach from \cite{kool2018attention}, we generate test sets of 10,000 instances for each problem class. The instances are sampled from the same distribution as the instances used in the experiments in \cite{kool2018attention} and \cite{nazari2018reinforcement}. We run the AM approach using the code and the models made available by the authors on the same machine as NLNS (sampling 1280 solutions for each instance). For the RL-BS method from \cite{nazari2018reinforcement}, we do not run any experiments ourselves (because no models are publicly available), but instead present the results reported by the authors.
These have been obtained using a test sets of 1,000 instances and a beam width of size 10.

Table~\ref{tab:exp_1} shows the average costs of all solutions for each instance set and the total wall-clock time for solving all instances of each instance set.
For the RL-BS approach, we assume that the runtime scales linearly with the size of the test set and show the runtime reported by the authors for 1,000 instances times 10 (and thus we mark these with a star). 

\begin{table}
\begin{center}
\caption{Comparison to existing machine learning based approaches.} \label{tab:exp_1}
\resizebox{\columnwidth}{!}{%
\begin{tabular}{lr|rrr|rrr}
&        &                         \multicolumn{3}{c|}{Avg. Costs} & \multicolumn{3}{|c}{Total Time (s)} \\  \hline
\multicolumn{2}{c|}{Instance Set}    & NLNS    & AM    & RL-BS   & NLNS    & AM   & RL-BS   \\ \hline
\parbox[t]{2mm}{\multirow{3}{*}{\rotatebox[origin=c]{90}{CVRP}}}
                       & $20$ customers  &  6.19    & 6.25  &   6.40      &  431    &  451   &  2*       \\
                       & $50$ customers  &  10.54       & 10.62 &    11.15     &  1453       &  1471  & 2*         \\
                       & $100$ customers&  15.99       & 16.20 &    16.96     &  3737      &  3750  & 4*        \\ \hline
\parbox[t]{2mm}{\multirow{3}{*}{\rotatebox[origin=c]{90}{SDVRP}}}
                       & $20$ customers  &  6.15       & 6.25  &  6.34   &   611    & 615   &   2*      \\
                       & $50$ customers  &  10.50     & 10.59 &   11.08    &   1934      & 1978   & 3*         \\
                       & $100$ customers &  16.00       & 16.27 &  16.86    &   5660      & 5691   & 5*        
\end{tabular}
}
\end{center}
\end{table}

NLNS significantly outperforms the AM and the RL-BS approach with respect to the solution costs on all instance sets. In comparison to the AM approach, NLNS finds better solutions in roughly the same amount of time. RL-BS finds significantly worse results than NLNS and the AM approach, but in a much shorter amount of time. Note that the AM approach can build solutions in a sequential manner, making it the method of choice when a solution to a single instance should be generated extremely quickly.

\subsection{Single Instance Search}
We next compare the single instance search of NLNS to state-of-the-art optimization approaches, and to an LNS using a handcrafted repair operator. We note that we do not tune the hyperparameters of NLNS and that the results can thus likely be improved. In all experiments we use a batch size of $300$ and we set $Z$ (the percentage of the batch that is reset to the current solution) to $0.8$. The starting temperature $T_s$ and the reheating temperature $T_r$ are dynamically set to the interquartile range of the costs of the solutions in $B$, and the cooling schedule exponentially reduces the temperature to $1$. We reheat the the temperature five times for instances with less than 200 customers, and ten times for instances with more customers. In all experiments we round the distances between the customers in the objective function to the nearest integer.

\subsubsection{Capacitated Vehicle Routing Problem}

For the CVRP we compare NLNS to the unified hybrid genetic search (UHGS) from \cite{vidal2012hybrid,vidal2014unified} and the heuristic solver LKH3 \cite{helsgaun2017extension}. UHGS is a population-based algorithm for VRP variants that consists of polymorphic components that can adapt to the problem type at hand. LKH3 is an extension to the Lin-Kernighan-Helsgaun TSP solver that is able to solve a variety of well-known routing problems. Furthermore, we evaluate the contribution of the learning-based components of NLNS. To this end, we replace the learned repair operators with the repair operator proposed in \cite{christiaens2016fresh}. The repair operator first sorts all customers that have been removed by the destroy operator based on one randomly selected criteria (e.g, customer demand). Customers are then reinserted into tours in a sequential fashion. For each customer, all potential insertion positions are evaluated and each position is selected with a certain probability depending on the costs of insertion and a parameter defining the greediness of the procedure. The hyperparameters of the destroy operator have  been selected similar to \cite{christiaens2016fresh}. Apart from the repair operation the learning-based NLNS and the non-learning-based LNS share the same components allowing for a direct comparison of the employed repair operations.

We evaluate all methods on a new dataset that contains instances that have the same properties as the instances proposed in \cite{uchoa2017new}. The instances from \cite{uchoa2017new} are all unique, but NLNS is meant to learn how to solve specific groups of instances. We thus organize the new dataset into 17 groups of 20 instances each with all instances of a group having the same characteristics (i.e., the instances have been sampled from the same distribution). This new dataset provides a middle ground between the datasets used in the machine learning literature (e.g., 10,000 instances with the same characteristic) and the optimization literature (e.g, a handful of instances all having different characteristics) and allows for a better evaluation of methods that are trained or tuned to solve instances of certain characteristic (such as machine learning based methods). 
The instance groups are shown in Table~\ref{tab:exp_2} and differ from each other in the following characteristics: the number of customers (\# Cust.), the location of the depot (Dep.), the position of the customers (Cust. Pos.), the demand of the customers (Dem.) and the capacity of the vehicle (Q). For a full description of the instance properties we refer readers to \cite{uchoa2017new}.

\begin{table*}
\setlength\tabcolsep{4pt}
\begin{center}
\caption{Comparison to state-of-the-art CVRP optimization approaches. Gap to UHGS in parentheses.} \label{tab:exp_2}
\resizebox{1.95\columnwidth}{!}{%
\begin{tabular}{l|rrrrr|rrrr|rrrr}
             & \multicolumn{5}{c|}{Instance Characteristics} & \multicolumn{4}{c|}{Avg. Costs}                                                                                          & \multicolumn{4}{c}{Avg. Time (s)} \\ \hline
Instance Set & \# Cust.  & Dep. & Cust. Pos. & Dem.   & Q    & NLNS                                        & LNS               & LKH3                                        & UHGS     & NLNS    & LNS    & LKH3   & UHGS   \\ \hline
XE\_1        & 100       & R    & RC(7)      & 1-100  & 206  & 30243.3 (\textbf{0.32\%})  & 30414.0 (0.89\%)  & 30785.0  (2.12\%)                           & 30146.4  & 191     & 192    & 372    & 36     \\
XE\_2        & 124       & R    & C(5)       & Q      & 188  & 36577.3  (\textbf{0.55\%}) & 36905.8 (1.45\%)  & 37226.9 (2.33\%)                            & 36378.3  & 191     & 192    & 444    & 64     \\
XE\_3        & 128       & E    & RC(8)      & 1-10   & 39   & 33584.6  (\textbf{0.44\%}) & 34122.5 (2.05\%)  & 33620.0  (0.54\%)                           & 33437.8  & 190     & 192    & 122    & 74     \\
XE\_4        & 161       & C    & RC(8)      & 50-100 & 1174 & 13977.5  (\textbf{0.72\%}) & 15002.5 (8.11\%)  & 13984.9  (0.78\%)                           & 13877.2  & 191     & 194    & 32     & 54     \\
XE\_5        & 180       & R    & C(6)       & U      & 8    & 26716.1 (0.58\%)                            & 27246.4 (2.58\%)  & 26604.4  (\textbf{0.16\%}) & 26562.4  & 191     & 193    & 65     & 86     \\
XE\_6        & 185       & R    & R          & 50-100 & 974  & 21700.0  (\textbf{1.09\%}) & 24071.2 (12.14\%)\hspace{-4pt} & 21705.2  (1.15\%)                           & 21465.7  & 191     & 195    & 100    & 101    \\
XE\_7        & 199       & R    & C(8)       & Q      & 402  & 29202.4 (2.03\%)                            & 30273.1 (5.78\%)  & 28824.9 (\textbf{0.72\%})  & 28620.0  & 191     & 195    & 215    & 142    \\
XE\_8        & 203       & C    & RC(6)      & 50-100 & 836  & 20593.0 (\textbf{0.51\%})  & 21038.2 (2.68\%)  & 20768.6  (1.37\%)                           & 20488.8  & 612     & 618    & 123    & 103    \\
XE\_9        & 213       & C    & C(4)       & 1-100  & 944  & 12218.7  (2.26\%)                           & 12828.4 (7.37\%)  & 12078.6 (\textbf{1.09\%})  & 11948.2  & 613     & 624    & 66     & 145    \\
XE\_10       & 218       & E    & R          & U      & 3    & 117642.1 (\textbf{0.04\%}) & 119750.8 (1.83\%) & 117699.8 (0.08\%)                           & 117600.1 & 612     & 616    & 112    & 138    \\
XE\_11       & 236       & E    & R          & U      & 18   & 27694.7 (\textbf{0.65\%})  & 30132.4 (9.50\%)  & 27731.5  (0.78\%)                           & 27517.0  & 613     & 621    & 67     & 190    \\
XE\_12       & 241       & E    & R          & 1-10   & 28   & 79413.6  (\textbf{3.10\%}) & 80383.3 (4.36\%)  & 79626.8  (3.38\%)                           & 77023.7  & 614     & 618    & 266    & 257    \\
XE\_13       & 269       & C    & RC(5)      & 50-100 & 585  & 34272.3 (\textbf{0.82\%})  & 35256.8 (3.71\%)  & 34521.5 (1.55\%)                            & 33994.7  & 613     & 618    & 343    & 214    \\
XE\_14       & 274       & R    & C(3)       & U      & 10   & 29032.5  (1.60\%)                           & 29168.0 (2.08\%)  & 28646.3 (\textbf{0.25\%})  & 28573.9  & 614     & 620    & 77     & 320    \\
XE\_15       & 279       & E    & R          & SL     & 192  & 45440.9 (1.81\%)                            & 47144.9 (5.63\%)  & 45224.6 (\textbf{1.32\%})  & 44633.5  & 615     & 629    & 347    & 329    \\
XE\_16       & 293       & C    & R          & 1-100  & 285  & 49259.3 (\textbf{0.57\%})  & 51080.8 (4.29\%)  & 50415.9 (2.93\%)                            & 48981.4  & 614     & 624    & 560    & 251    \\
XE\_17       & 297       & R    & R          & 1-100  & 55   & 37096.3 (1.41\%)                            & 39324.0 (7.50\%)  & 37031.9 (\textbf{1.23\%})  & 36582.1  & 615     & 623    & 152    & 250   
\end{tabular}
}
\end{center}
\end{table*}

We run NLNS using four repair operators (corresponding to four destroy operators with different destroy procedures and degrees of destruction) that are trained separately for each instance group on separate training instances. LKH3 is evaluated on a single core of an Intel Xeon E5-2670 CPU at 2.6 GHz with the hyperparameter configuration used by \cite{helsgaun2017extension} to solve the instances of \cite{uchoa2017new}. Each method is run three times for each instance. Table~\ref{tab:exp_2} reports the average costs and the average runtime for each instance group. We do not report the gap to optimality because it not computationally feasibly to compute optimal solutions for all instances. UHGS offers the best performance on all instance groups, albeit with only small gaps between 0.08\% to 3.38\% to NLNS and LKH3.  The LNS using a handcrafted heuristic is significantly outperformed by NLNS on all instances with gaps between 0.89\% to 12.14\% to UHGS. While this does not show that designing a repair operator that can compete with NLNS is impossible, it indicates that this is at best a complex task that requires significant effort and a deep understanding of the problem.
NLNS and LKH3 offer similar performance with NLNS finding solutions with lower costs on 11 of 17 instance groups. NLNS is the first learned heuristic to achieve performance parity with LKH3, which is a significant achievement given the many years of human development and insight that have gone into LKH3.

\subsubsection{Split Delivery Vehicle Routing Problem}
We compare NLNS to the state-of-the-art multi-start iterated local search algorithm called SplitILS from \cite{silva2015iterated} on eight SDVRP instances from \cite{belenguer2000lower}. The instances have either 75 or 100 customers, and the demands of the customers are chosen to be within a different interval for each instance. The vehicle capacity $Q$ is 160.

For NLNS we train only two sets of repair operators each consisting of 4 repair operators (with different destroy procedures and degrees of destruction). Both sets of repair operators are trained on instances with 100 randomly positioned customers, but for the first set customers have a demand randomly selected in $[0.01Q, 0.3Q]$ and for the second set in $[0.1Q, 0.9Q]$.
We use the first set to solve the instances S76D1, S76D2, S101D1 and S101D2 and the second set to solve the rest of the instances. Note that repair operators are applied to instances that have been sampled from a distribution that does not exactly match the underlying distribution of the instances used for their training (e.g., repair operators trained on instances with 100 customers are used to repair instances with 75 customers). Table~\ref{tab:exp_3} shows the average performance (over 20 runs) for NLNS and Split\-ILS. While SplitILS finds better solutions than NLNS on instances with 75 customers, NLNS outperforms Split\-ILS on larger instances with 100 customers. Thus, as in the case of the CVRP, NLNS is able to match, and even slightly outperform, a state-of-the-art technique from the literature.

\begin{table}
\begin{center}
\caption{Comparison of NLNS and SplitILS on SDVRP instances.} \label{tab:exp_3}
\resizebox{0.9\columnwidth}{!}{%
\begin{tabular}{l|rr|rr|rr}
         & \multicolumn{2}{c|}{Demand}                            & \multicolumn{2}{c|}{Avg. Costs}                        & \multicolumn{2}{c}{Avg. Time (s)} \\ \hline
Instance & min                      & max                       & \multicolumn{1}{l}{NLNS} & \multicolumn{1}{l|}{SplitILS} & NLNS            & SplitILS            \\ \hline
S76D1    & 0.01Q                    & 0.1Q                      & 593.00                   & \textbf{592.45}            & 191             & 5                \\
S76D2    & 0.1Q                     & 0.3Q                      & 1085.60                  & \textbf{1083.35}           & 190             & 59               \\
S76D3    & 0.1Q                     & 0.5Q                      & 1424.45                  & \textbf{1422.05}           & 190             & 8                \\
S76D4    & 0.1Q                     & 0.9Q                      & 2075.50                  & \textbf{2074.30}           & 190             & 148              \\
S101D1   & 0.01Q                    & 0.1Q                      & \textbf{718.30}          & 718.40                     & 311             & 14               \\
S101D2   & 0.1Q                     & 0.3Q                      & \textbf{1367.35}         & 1370.95                    & 311             & 116              \\
S101D3   & 0.1Q & 0.5Q & \textbf{1865.70}         & 1868.75                    & 311             & 233              \\
S101D5   & 0.7Q & 0.9Q & \textbf{2774.50}         & 2779.65                    & 311             & 580             
\end{tabular}
}
\end{center}
\end{table}

\section{CONCLUSION}
We presented NLNS, an extension to LNS that learns repair operators for VRP instances and employs them in a guided heuristic search. We compared NLNS on CVRP and SDVRP instances to state-of-the-art machine learning and optimization approaches and show that the learned heuristic components of NLNS outperform a handcrafted heuristic from the literature. The search guidance of NLNS also allows it to significantly outperform machine learning based approaches (relying on trivial search strategies) when solving a set of instances in parallel. When solving CVRP instances sequentially, NLNS offers similar performance to the heuristic solver LKH3, but is outperformed by the population-based UHGS approach. On SDVRP instances, NLNS is able to outperform the state-of-the-art method SplitILS on instances with 100 customers. Our results convincingly show that combining learned models with high-level search guidance outperform models lacking search capabilities. In future work, we will further explore the relationship between learned models and search procedures, including making joint training of repair and destroy operators possible.

\ack
We thank Thibaut Vidal for running the UHGS method on our CVRP dataset. The computational results in this work are performed using the Bielefeld GPU Cluster. We also thank Yuri Malitsky for providing helpful feedback on this paper.

\bibliography{bib}

\begin{thebibliography}{10}

\bibitem{ansotegui2018hyper}
C.~Ans{\'o}tegui, B.~Heymann, J.~Pon, M.~Sellmann, and K.~Tierney.
\newblock Hyper-reactive tabu search for maxsat.
\newblock In {\em International Conference on Learning and Intelligent
  Optimization}, pages 309--325. Springer, 2018.

\bibitem{belenguer2000lower}
J.-M. Belenguer, M.~Martinez, and E.~Mota.
\newblock A lower bound for the split delivery vehicle routing problem.
\newblock {\em Operations research}, 48(5):801--810, 2000.

\bibitem{bello2016neural}
I.~Bello, H.~Pham, Q.~V. Le, M.~Norouzi, and S.~Bengio.
\newblock Neural combinatorial optimization with reinforcement learning.
\newblock {\em arXiv preprint arXiv:1611.09940}, 2016.

\bibitem{bengio2018machine}
Y.~Bengio, A.~Lodi, and A.~Prouvost.
\newblock Machine learning for combinatorial optimization: a methodological
  tour d'horizon.
\newblock {\em arXiv preprint arXiv:1811.06128}, 2018.

\bibitem{bresson2017residual}
X.~Bresson and T.~Laurent.
\newblock Residual gated graph convnets.
\newblock {\em arXiv preprint arXiv:1711.07553}, 2017.

\bibitem{christiaens2016fresh}
J.~Christiaens and G.~Vanden~Berghe.
\newblock A fresh ruin \& recreate implementation for the capacitated vehicle
  routing problem.
\newblock 2016.

\bibitem{dantzig1959truck}
G.~B. Dantzig and J.~H. Ramser.
\newblock The truck dispatching problem.
\newblock {\em Management science}, 6(1):80--91, 1959.

\bibitem{deudon2018learning}
M.~Deudon, P.~Cournut, A.~Lacoste, Y.~Adulyasak, and L.-M. Rousseau.
\newblock Learning heuristics for the {TSP} by policy gradient.
\newblock In {\em International Conference on the Integration of Constraint
  Programming, Artificial Intelligence, and Operations Research}, pages
  170--181. Springer, 2018.

\bibitem{helsgaun2017extension}
K.~Helsgaun.
\newblock An extension of the {Lin-Kernighan-Helsgaun TSP} solver for
  constrained traveling salesman and vehicle routing problems.
\newblock {\em Roskilde: Roskilde University}, 2017.

\bibitem{hopfield1985neural}
J.~J. Hopfield and D.~W. Tank.
\newblock {N}eural computation of decisions in optimization problems.
\newblock {\em Biological cybernetics}, 52(3):141--152, 1985.

\bibitem{DLTS}
A.~Hottung, S.~Tanaka, and K.~Tierney.
\newblock Deep learning assisted heuristic tree search for the container
  pre-marshalling problem.
\newblock {\em Computers \& Operations Research}, 113:104781, 2020.

\bibitem{joshi2019efficient}
C.~K. Joshi, T.~Laurent, and X.~Bresson.
\newblock An efficient graph convolutional network technique for the travelling
  salesman problem.
\newblock {\em arXiv preprint arXiv:1906.01227}, 2019.

\bibitem{kaempfer2018learning}
Y.~Kaempfer and L.~Wolf.
\newblock Learning the multiple traveling salesmen problem with permutation
  invariant pooling networks.
\newblock {\em arXiv preprint arXiv:1803.09621}, 2018.

\bibitem{khalil2017learning}
E.~Khalil, H.~Dai, Y.~Zhang, B.~Dilkina, and L.~Song.
\newblock Learning combinatorial optimization algorithms over graphs.
\newblock In {\em Advances in Neural Information Processing Systems}, pages
  6348--6358, 2017.

\bibitem{kingma2014adam}
D.~Kingma and J.~Ba.
\newblock Adam: A method for stochastic optimization.
\newblock {\em arXiv preprint arXiv:1412.6980}, 2014.

\bibitem{kirkpatrick1983optimization}
S.~Kirkpatrick, C.~D. Gelatt, and M.~P. Vecchi.
\newblock Optimization by simulated annealing.
\newblock {\em Science}, 220(4598):671--680, 1983.

\bibitem{kool2018attention}
W.~Kool, H.~van Hoof, and M.~Welling.
\newblock Attention, learn to solve routing problems!
\newblock In {\em International Conference on Learning Representations}, 2019.

\bibitem{kotthoff2016algorithm}
L.~Kotthoff.
\newblock Algorithm selection for combinatorial search problems: A survey.
\newblock In {\em Data Mining and Constraint Programming}, pages 149--190.
  Springer, 2016.

\bibitem{nazari2018reinforcement}
M.~Nazari, A.~Oroojlooy, L.~Snyder, and M.~Tak{\'a}c.
\newblock Reinforcement learning for solving the vehicle routing problem.
\newblock In {\em Advances in Neural Information Processing Systems}, pages
  9839--9849, 2018.

\bibitem{nowak2017note}
A.~Nowak, S.~Villar, A.~S. Bandeira, and J.~Bruna.
\newblock A note on learning algorithms for quadratic assignment with graph
  neural networks.
\newblock {\em Stat}, 1050:22, 2017.

\bibitem{ropke2006adaptive}
S.~Ropke and D.~Pisinger.
\newblock An adaptive large neighborhood search heuristic for the pickup and
  delivery problem with time windows.
\newblock {\em Transportation Science}, 40(4):455--472, 2006.

\bibitem{shaw1998using}
P.~Shaw.
\newblock Using constraint programming and local search methods to solve
  vehicle routing problems.
\newblock In {\em Fourth International Conference on Principles and Practice of
  Constraint Programming}, pages 417--431. Springer, 1998.

\bibitem{silva2015iterated}
M.~M. Silva, A.~Subramanian, and L.~S. Ochi.
\newblock An iterated local search heuristic for the split delivery vehicle
  routing problem.
\newblock {\em Computers \& Operations Research}, 53:234--249, 2015.

\bibitem{sutskever2014sequence}
I.~Sutskever, O.~Vinyals, and Q.~V. Le.
\newblock Sequence to sequence learning with neural networks.
\newblock In {\em Advances in Neural Information Processing Systems}, pages
  3104--3112, 2014.

\bibitem{syed2019neural}
A.~A. Syed, K.~Akhnoukh, B.~Kaltenhaeuser, and K.~Bogenberger.
\newblock Neural network based large neighborhood search algorithm for ride
  hailing services.
\newblock In {\em EPIA Conference on Artificial Intelligence}, pages 584--595.
  Springer, 2019.

\bibitem{tyasnurita2017learning}
R.~Tyasnurita, E.~{\"O}zcan, and R.~John.
\newblock Learning heuristic selection using a time delay neural network for
  open vehicle routing.
\newblock In {\em 2017 IEEE Congress on Evolutionary Computation (CEC)}, pages
  1474--1481. IEEE, 2017.

\bibitem{uchoa2017new}
E.~Uchoa, D.~Pecin, A.~Pessoa, M.~Poggi, T.~Vidal, and A.~Subramanian.
\newblock New benchmark instances for the capacitated vehicle routing problem.
\newblock {\em European Journal of Operational Research}, 257(3):845--858,
  2017.

\bibitem{velivckovic2017graph}
P.~Veli{\v{c}}kovi{\'c}, G.~Cucurull, A.~Casanova, A.~Romero, P.~Lio, and
  Y.~Bengio.
\newblock Graph attention networks.
\newblock {\em arXiv preprint arXiv:1710.10903}, 2017.

\bibitem{vidal2012hybrid}
T.~Vidal, T.~G. Crainic, M.~Gendreau, N.~Lahrichi, and W.~Rei.
\newblock A hybrid genetic algorithm for multidepot and periodic vehicle
  routing problems.
\newblock {\em Operations Research}, 60(3):611--624, 2012.

\bibitem{vidal2014unified}
T.~Vidal, T.~G. Crainic, M.~Gendreau, and C.~Prins.
\newblock A unified solution framework for multi-attribute vehicle routing
  problems.
\newblock {\em European Journal of Operational Research}, 234(3):658--673,
  2014.

\bibitem{vinyals2015pointer}
O.~Vinyals, M.~Fortunato, and N.~Jaitly.
\newblock Pointer networks.
\newblock In {\em Advances in Neural Information Processing Systems}, pages
  2692--2700, 2015.

\bibitem{williams1992simple}
R.~J. Williams.
\newblock Simple statistical gradient-following algorithms for connectionist
  reinforcement learning.
\newblock {\em Machine Learning}, 8(3-4):229--256, 1992.

\end{thebibliography}
\end{document}